\documentclass[10pt,twocolumn,letterpaper]{article}

\usepackage[hyphens]{url} 
\usepackage{makecell}

\usepackage{epsfig}
\usepackage{graphicx}
\usepackage{amsmath}
\usepackage{amssymb}
\usepackage{booktabs}
\usepackage[ruled,vlined]{algorithm2e}
\usepackage{algorithmic}
\usepackage[dvipsnames]{xcolor} 
\usepackage{colortbl} 
\usepackage{multirow}
\usepackage{arydshln}
\usepackage{tikz}
\usetikzlibrary{tikzmark}
\usepackage{siunitx} 
\usepackage{mdframed}
\usepackage{tablefootnote} 

\usepackage{fontawesome}
\newcommand{\huggingface}{\raisebox{-1.5pt}{\includegraphics[height=1.05em]{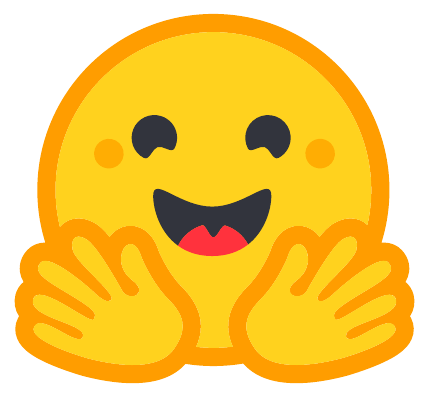}}\xspace}
\newcommand{\github}{\raisebox{-1.5pt}{\includegraphics[height=1.05em]{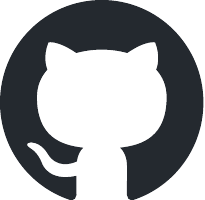}}\xspace}

\newcommand{\worldwideweb}{\raisebox{-1.5pt}{\includegraphics[height=1.05em]{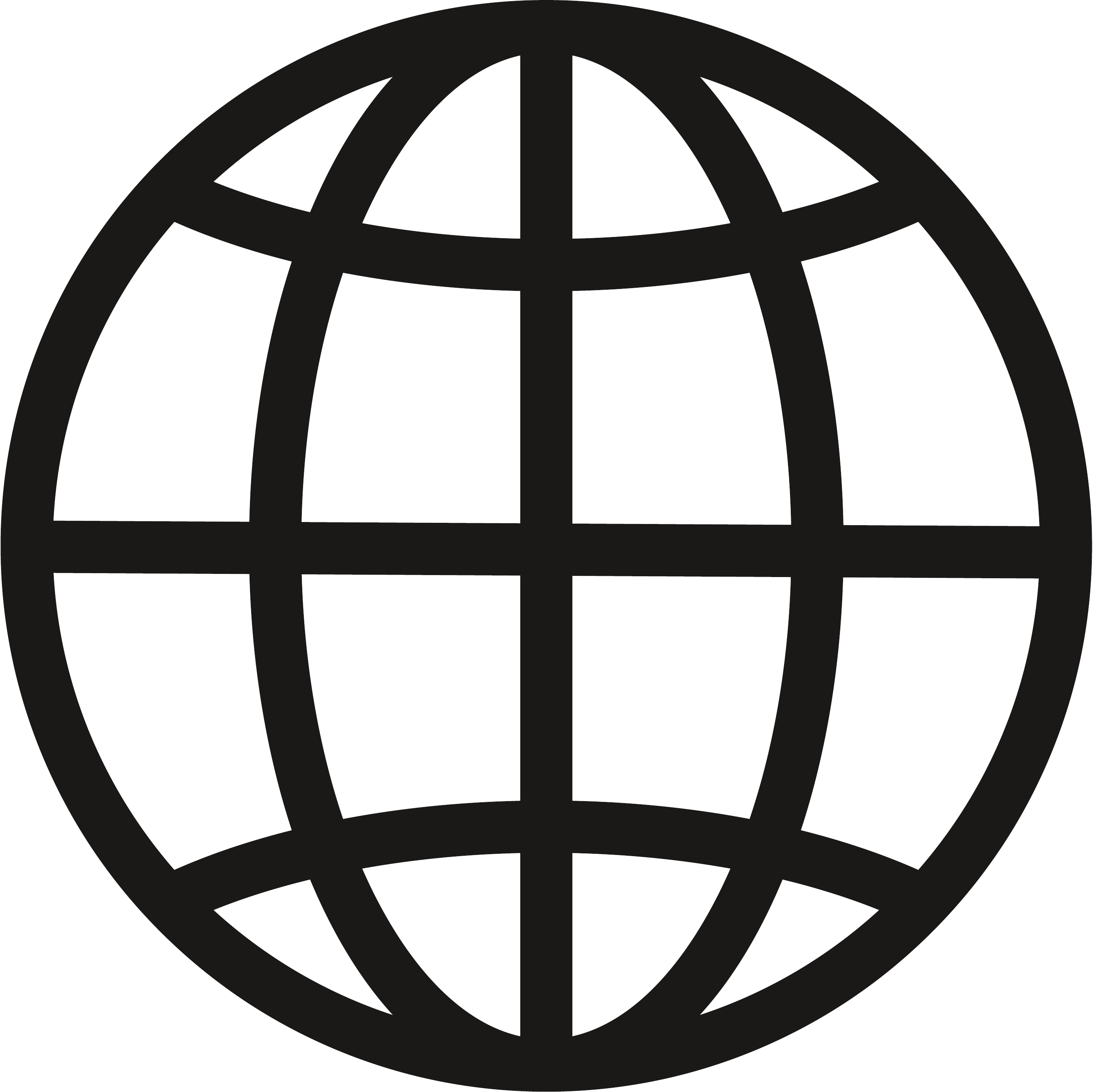}}\xspace}

\newcolumntype{g}{>{\columncolor{gray!30}}c}

\usepackage{tcolorbox}
\tcbuselibrary{most}

\definecolor{citecolor}{HTML}{0071bc}

\tcbuselibrary{skins}

\definecolor{userbg}{RGB}{245, 245, 245}
\definecolor{userborder}{RGB}{210, 229, 255}
\definecolor{userfont}{RGB}{0, 0, 0}

\tcbset{
  enhanced,
  boxrule=1pt,
  colback=userbg,
  colframe=userborder,
  fonttitle=\bfseries,
  coltitle=userfont,
  boxsep=5pt,
  arc=5pt,
  outer arc=5pt,
  left=10pt,
  right=10pt,
  top=2pt,
  bottom=2pt,
  attach boxed title to top left={yshift=-0.25\baselineskip-1mm,xshift=2mm},
  boxed title style={
    colback=userborder,
    sharp corners,
    rounded corners=northeast,
    arc=3pt,
    outer arc=3pt,
    boxrule=0pt,
    boxsep=2pt,
  },
}
\usepackage{pgf}
\usepackage{adjustbox}
\usepackage[shortlabels,inline]{enumitem} 
\usepackage{array}
\definecolor{listcolor}{RGB}{50,120,230}
\usepackage{caption}

\usepackage{titlesec}
\titlespacing*{\paragraph}
    {0pt}     
    {0.3em}   
    {0.5em}     

\iftrue   
    \newcommand{\displaytodo}[1]{#1}
\else
    \newcommand{\displaytodo}[1]{}
\fi

\newcommand{\DDT}{$\text{DiT}^{\text{DH}}$\xspace} 

\newcounter{researchquestion}

\newcommand{\researchquestion}[2][]{%
    \vspace{0.8em}
    \refstepcounter{researchquestion}
    \begin{tcolorbox}[
        enhanced,
        colback=blue!5,
        colframe=blue!70!black,
        fonttitle={\fontsize{10.5pt}{12.8pt}\selectfont\bfseries\color{blue!20!black}},  
        title=Question \theresearchquestion,
        toprule=1.5pt,
        bottomrule=0.8pt,
        leftrule=0.8pt,
        rightrule=0.8pt,
        left=6pt,
        right=6pt,
        top=6pt,
        bottom=6pt,
        boxsep=3pt
    ]
    \normalsize #2
    \end{tcolorbox}
    \ifx\\#1\\\else\label{rq:#1}\fi
    \vspace{0.5em}
}

\usepackage[pagenumbers]{cvpr} 

\definecolor{cvprblue}{rgb}{0.21,0.49,0.74}
\usepackage[pagebackref,breaklinks,colorlinks,allcolors=cvprblue]{hyperref}

\def\paperID{18873} 
\def\confName{CVPR}
\def\confYear{2026}
\newcommand{\linkbutton}[3]{%
    \href{#1}{\tcbox[on line, arc=3pt, colback=white, colframe=white, boxrule=0.5pt, size=small, left=1mm, right=1mm, top=0.5mm, bottom=0.5mm, fontupper=\small\sffamily]{#2 \textbf{#3}}}%
}

\title{Scaling Text-to-Image Diffusion Transformers with Representation Autoencoders}

\author{Shengbang Tong\textsuperscript{*}, Boyang Zheng\textsuperscript{*}, Ziteng Wang\textsuperscript{*}, Bingda Tang, Nanye Ma,\\ Ellis Brown, Jihan Yang, Rob Fergus, Yann LeCun, Saining Xie\\
New York University\\
\linkbutton{https://rae-dit.github.io/scale-rae/}{\worldwideweb}{Website}\hspace{4mm}%
\linkbutton{https://github.com/ZitengWangNYU/Scale-RAE}{\github}{Code}\hspace{4mm}%
\linkbutton{https://huggingface.co/collections/nyu-visionx/scale-rae}{\huggingface}{Models}\hspace{4mm}%
\linkbutton{https://huggingface.co/datasets/nyu-visionx/scale-rae-data}{\huggingface}{Data}
}
\begin{document}
\maketitle
\def\thefootnote{*}\footnotetext{Core contributor.}\def\thefootnote{\arabic{footnote}}

\begin{abstract}
    Representation Autoencoders (RAEs) have shown distinct advantages in diffusion modeling on ImageNet by training in high-dimensional semantic latent spaces.
    In this work, we investigate whether this framework can scale to large-scale, freeform text-to-image (T2I) generation.
    We first scale RAE decoders on the frozen representation encoder (SigLIP-2) beyond ImageNet by training on web, synthetic, and text-rendering data, finding that while scale improves general fidelity, targeted data composition is essential for specific domains like text.
We then rigorously stress-test the RAE design choices originally proposed for ImageNet. Our analysis reveals that scaling simplifies the framework: while dimension-dependent noise scheduling remains critical, architectural complexities such as wide diffusion heads and noise-augmented decoding offer negligible benefits at scale
Building on this simplified framework, we conduct a controlled comparison of RAE against the state-of-the-art FLUX VAE across diffusion transformer scales from 0.5B to 9.8B parameters.
RAEs consistently outperform VAEs during pretraining across all model scales. Further, during finetuning on high-quality datasets, VAE-based models catastrophically overfit after 64 epochs, while RAE models remain stable through 256 epochs and achieve consistently better performance.
Across all experiments, RAE-based diffusion models demonstrate faster convergence and better generation quality, establishing RAEs as a simpler and stronger foundation than VAEs for large-scale T2I generation.
    Additionally, because both visual understanding and generation can operate in a shared representation space, the multimodal model can directly reason over generated latents, opening new possibilities for unified models.
\end{abstract}

\section{Introduction}\label{sec:intro}
\begin{figure}[t]
    \centering
    \includegraphics[width=\columnwidth]{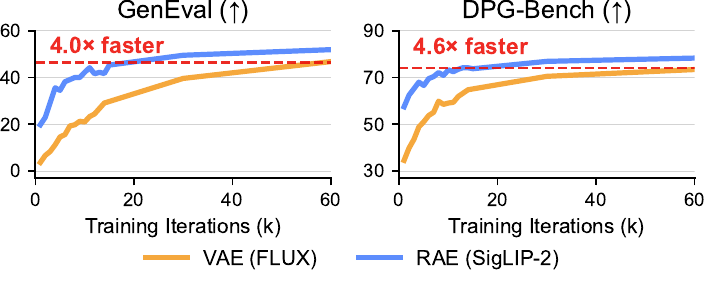}
    \caption{
   \textbf{RAE converges faster than VAE in text-to-image pretraining.}
    We train Qwen-2.5 1.5B + DiT 2.4B models from scratch on both RAE (SigLIP-2) and VAE (FLUX) latent spaces for up to 60k iterations.
    RAE converges significantly faster than VAE on both GenEval (4.0×) and DPG-Bench (4.6×).
    }\label{fig:convergence}
\end{figure}

\begin{figure*}[t]
  \centering
    \centering
    \includegraphics[width=\linewidth]{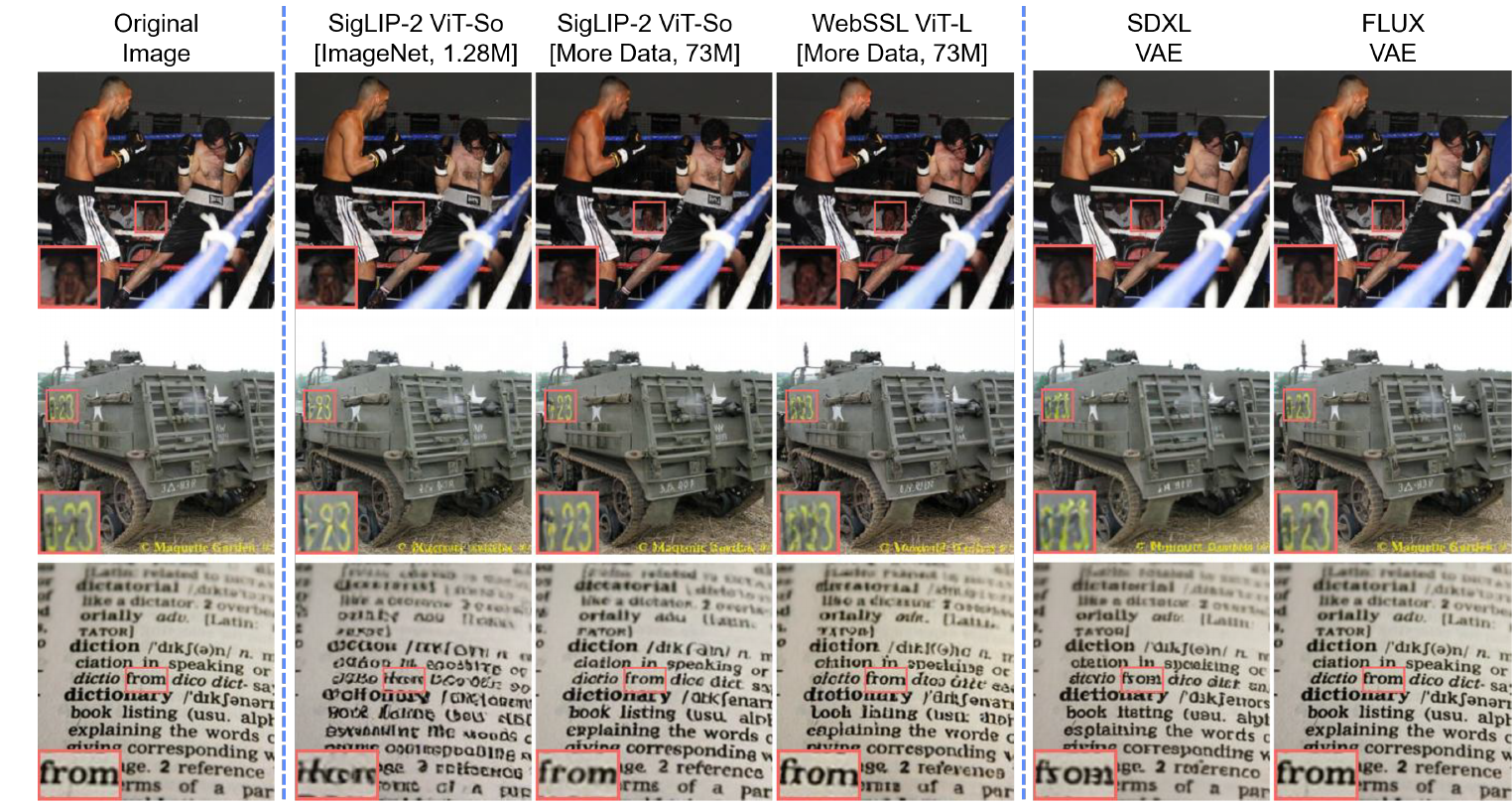}
    \caption{
        \textbf{RAE decoders trained on more data (web, synthetic \& text) generalize across domains.} 
        Decoders trained only on ImageNet reconstruct natural images well but struggle with text-rendering scenes (see second column). Adding web and text data greatly improves text reconstruction while maintaining natural-image quality. We also observe that both the language-supervised model and the SSL model learn representations suitable for reconstructing diverse images, including natural languages. Compared to proprietary VAEs, our RAE models achieve competitive overall fidelity.
        }
    \label{fig:comparison}
\end{figure*}

Diffusion-based generative modeling~\citep{adm, ddpm, fm} has made rapid progress, giving rise to state-of-the-art systems across visual generative domains such as text-to-image generation~\citep{StabilityAI2024SD35, flux, wu2025qwen}. A key factor in this success is the adoption of \emph{latent diffusion}~\citep{Rombach_2022_CVPR}, where generation occurs in a compact latent space encoded by a variational autoencoder (VAE)~\citep{VAE}, rather than directly in pixel space. 

In parallel with advances in generative modeling, visual representation learning has progressed through self-supervised learning (SSL)~\citep{simclr, moco, MAE, DINO}, language supervision~\citep{radford2021learning, zhai2023sigmoid}, and their combinations~\citep{mu2021slip, siglip2}. These models produce semantically structured, high-dimensional representations that generalize well across visual understanding tasks.
Unlike VAE encoders, which compress images into \textit{low-dimensional} latents, the representation encoders operate on \emph{high-dimensional} latents that can capture much more semantically rich features.

Such high-dimensional latents were previously considered too ``abstract'' for effective generative modeling~\citep{ImprovDiffus,lgt}, or outright intractable~\citep{pgv3, vugen}. However, a recent approach, Representation Autoencoder (RAE)~\citep{zheng2025diffusion}, has paved a path forward by training decoders on frozen representation encoders.
RAE pairs a powerful frozen representation encoder with a lightweight trained decoder to reconstruct pixels from high-dimensional embeddings, enabling diffusion directly in this semantic latent space.
In the highly controlled class-conditional ImageNet~\citep{deng2009imagenet} setting, RAE demonstrates that diffusion in such frozen representation spaces can achieve more efficient and effective training than conventional VAE-based diffusion.

However, ImageNet represents a best-case scenario: fixed resolution, curated content, and class-conditional generation.
A critical question remains unanswered: \emph{can RAE truly scale to the complexities of freeform text-to-image generation?}
This setting involves broader visual diversity, open-ended compositions, and substantially larger models and compute---challenges for which high-dimensional latent diffusion remains unproven.

In this work, we investigate whether RAEs can succeed at scale by training diffusion models for large-scale text-to-image (T2I) generation. We adopt SigLIP-2~\citep{tschannen2025siglip} as the frozen representation encoder and use the MetaQuery framework~\citep{metaquery} to train a unified T2I model, leveraging a powerful pretrained large language model (LLM)~\citep{qwen2024qwen2}.

As a first step, we study decoder training beyond ImageNet supervision (\cref{sec: scale decoder}). Expanding from ImageNet to web-scale and synthetic aesthetic data yields only \textit{small} gains on ImageNet itself, but provides \textit{moderate} improvements on more diverse natural images such as YFCC~\citep{thomee2016yfcc100m}, showing that broader distributions enhance generalization. However, we find that text reconstruction requires targeted supervision: without text-specific data, the decoder fails to reproduce fine glyph details. Adding text-rendering data leads to substantial improvements, highlighting that data \emph{composition}, not just scale, is crucial.

Next, we analyze design choices in the RAE framework~\citep{zheng2025diffusion} and evaluate their importance under large-scale T2I training (\cref{sec: revisiting rae}).  We find that scale acts as a simplifier.
Dimension-aware noise scheduling remains essential: removing the shift leads to substantially worse performance. The \emph{wide DDT head} (\DDT{}) provides clear benefits for smaller backbones, but its advantage fades as Diffusion Transformers (DiT) scale to the billion-parameters.
Finally, the effect of \emph{noise-augemented decoding} is modest at T2I scale, with gains saturating quickly.

We then systematically compare RAEs with \emph{SOTA} VAEs under matched training conditions (\cref{sec: vae vs rae}).
We train DiTs from \textbf{scratch} following the conventional two-stage T2I setup~\citep{dai2023emu,podell2023sdxl}:
(i) large-scale pretraining with randomly initialized DiTs, and
(ii) finetuning on smaller high-quality datasets.
During pretraining, RAE-based models converge significantly faster and achieve higher performance on both GenEval and DPG-Bench.
As shown in \cref{fig:convergence}, training a 1.5B LLM + 2.4B DiT with RAE (SigLIP-2) achieves a 4.0× speedup on GenEval and a 4.6× speedup on DPG-Bench compared to its VAE counterpart.
This advantage is consistent across both language backbones (Qwen-2.5~\cite{qwen} 1.5B--7B) and diffusion scales (DiT 0.5B--9.8B).
In finetuning, RAE models continue to outperform their VAE counterparts and are less prone to overfitting. 

Finally, we examine unified models in which RAE enables understanding and generation to operate in the same high-dimensional semantic space (\cref{sec: unified model implications}). We find that adding generative training does \emph{not} degrade understanding performance, and the choice of RAE vs.\ VAE in the generative path has little effect because both rely on the same frozen understanding encoder.
Moreover, the shared latent space allows the LLM to process generated latents directly, without decoding back to pixels. We take a first exploratory step toward leveraging this property through latent-space test-time scaling, which proves both feasible and effective.

Ultimately, we aim to convey one primary message:
\textbf{Representation Autoencoders provide a simpler and stronger foundation than VAEs for training large-scale text-to-image diffusion models.} 
They offer a simple yet effective path to scaling generation within semantic representation spaces. 
We will release \emph{all} code, data, and model checkpoints related to this work to foster open and reproducible research in multimodal generation.

\begin{table}[t]
    \centering
    \small
    \caption{
        \textbf{Data matters for RAE's reconstruction fidelity.}
        We train RAE (SigLIP-2) on different data sources. Compared with ImageNet-only training, using web-scale images consistently improves reconstruction quality across all domains.
    }\label{tab:rfid_vs_data}
    \setlength{\tabcolsep}{2.9pt}
    \begin{tabular}{lcccc}
        \toprule
        \textbf{Data Sources} & \textbf{\#Data} &\textbf{ImageNet} $\downarrow$ & \textbf{YFCC} $\downarrow$ & \textbf{Text} $\downarrow$ \\
        \midrule
        \rowcolor{gray!10}
        ImageNet  & 1.28M &0.462 & 0.970 & 2.640 \\
        \midrule
        Web & 39.3M  & 0.529 &\textbf{0.629} & 2.325 \\
        Web + Synthetic &64.0M & 0.437 & 0.683 & 2.406 \\
        \rowcolor{green!10}
        Web + Synthetic + Text& 73.0M & \textbf{0.435} &0.702 & \textbf{1.621} \\
        \bottomrule
    \end{tabular}
\end{table}

\section{Scaling Decoder Training Beyond ImageNet}\label{sec: scale decoder}

To adapt the RAE framework for open-world T2I generation, we first train a RAE decoder on a larger and more diverse dataset than ImageNet~\citep{deng2009imagenet}.
Throughout this section, we choose SigLIP-2 So400M (patch size 14)~\citep{tschannen2025siglip} as the frozen encoder, and train a ViT-based~\citep{dosovitskiy2020image} decoder to reconstruct images from these tokens at $224\times224$ resolution. We present the architectural details in \cref{appendix: implementation}. 
Given an input image $x \in \mathbb{R}^{3\times224\times224}$, the encoder produces $N = 16 \times 16$ tokens with channel dimension $d = 1152$.

\paragraph{Training objective.}
Following RAE, we adopt $\ell_1$, LPIPS~\citep{lpips}, and adversarial losses~\citep{styleganxl, GAN}. Additionally, we integrate Gram Loss~\citep{gramloss}, which is found beneficial for reconstruction~\citep{atoken}. The training objective is set as $
L(x, \hat{x}) = \ell_1(x, \hat{x}) + \omega_L \text{LPIPS}(x, \hat{x}) + \omega_G \text{Gram}(x, \hat{x}) +  \omega_A \text{Adv}(x,\hat{x}), \hat{x} = \text{RAE}(x)$. We include the weights and training details in \cref{appendix: implementation}.

\paragraph{Training data.}
We use a dataset combining roughly 73M data from three data sources: web image sources from FuseDiT~\citep{Tang_2025_CVPR}, synthetic images generated by FLUX.1-schnell~\citep{flux}, and RenderedText~\citep{wendlerc2024renderedtext}, which focuses on text-rendering scenes.
Details are provided in \cref{sec: vae vs rae}.

\paragraph{Evaluation.}
We evaluate rFID-50k~\citep{fid} of reconstructed images in three representative domains: 
(i) ImageNet-1k~\citep{imagenet} for classic object-centric evaluation, 
(ii) YFCC~\citep{thomee2016yfcc100m} for diverse web-scale imagery, 
and (iii) RenderedText~\citep{wendlerc2024renderedtext} held-out set for text-rendering and typography-specific evaluation. We evaluate rFiD on 50k samples from each data source and present our results in \cref{tab:rfid_vs_data,tab:rfid_vs_vae}.

\paragraph{Web-scale training of RAE decoders.}
As shown in \cref{tab:rfid_vs_data}, expanding decoder training beyond ImageNet to include web-scale and synthetic data yields only marginal gains on ImageNet itself, but provides moderate improvements on more diverse images (YFCC). This indicates that exposure to a broader distribution enhances the decoder’s generalizability. However, generic web data is insufficient for text reconstruction. Training on Web + Synthetic data yields little improvement over ImageNet-only training. In contrast, performance improves substantially once text-specific data is included, highlighting that reconstruction quality is very sensitive to the composition of the training data. As shown in~\cref{fig:comparison}, training the RAE decoder with additional text data is essential for accurate text reconstruction.
Overall, RAE reconstruction improves with scale, but the composition of data---not just its size---matters: each domain benefits most from domain-matched coverage.

\begin{table}[t]
    \centering
    \small
    \setlength{\tabcolsep}{5.2pt}  
    \caption{
        \textbf{Comparison of reconstruction performance.}
        After expanding the training data, RAE outperforms SDXL-VAE across all three domains, though it still falls short of FLUX-VAE. Within RAE variants, WebSSL reconstructs better than SigLIP-2.
    }\label{tab:rfid_vs_vae}
    \begin{tabular}{l l ccc}
        \toprule
        \textbf{Family} & \textbf{Model} & \textbf{ImageNet} $\downarrow$ & \textbf{YFCC} $\downarrow$ & \textbf{Text} $\downarrow$ \\
        \midrule
        \multirow{2}{*}{VAE}
            & SDXL & 0.930 & 1.168 & 2.057 \\
            & FLUX & 0.288 & 0.410 & 0.638 \\
        \midrule
        \multirow{2}{*}{RAE}
            & WebSSL ViT-L & 0.388 & 0.558 & 1.372 \\
            & \cellcolor{green!10}SigLIP-2 ViT-So 
                & \cellcolor{green!10}0.435 
                & \cellcolor{green!10}0.702 
                & \cellcolor{green!10}1.621 \\
        \bottomrule
    \end{tabular}
\end{table}

\begin{figure*}
    \centering
    \includegraphics[width=\linewidth]{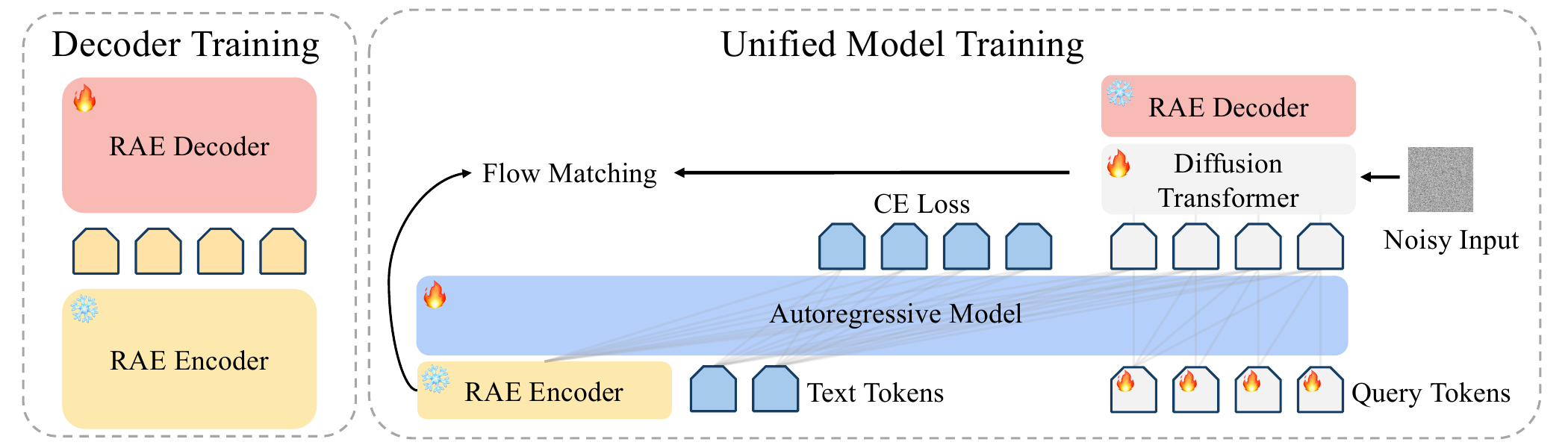}
    \caption{\textbf{Overview of training pipeline}. \textit{Left}: RAE decoder training stage. We train a decoder on the representations (yellow tokens) produced by the frozen RAE encoder. \textit{Right}: End-to-end unified training of the autoregressive model, diffusion transformer, and learnable query tokens (gray tokens) using cross-entropy (CE) loss for text prediction and a flow-matching objective for image prediction. }
    \label{fig:architecture}
    \vspace{-1em}
\end{figure*}

\paragraph{Different encoders.}
We also experiment training RAE using different pretrained encoders. In particular, we replace SigLIP-2 with WebSSL-DINO~\citep{fan2025scaling}, a large-scale self-supervised model. As shown in~\cref{tab:rfid_vs_vae}, WebSSL-DINO achieves stronger reconstruction performance than SigLIP-2 across all domains, including text reconstruciton. Both SigLIP-2 and WebSSL-L consistently outperform SDXL VAE~\citep{podell2023sdxl}, though they still fall short of FLUX VAE~\citep{flux}.

\section{RAE is Simpler in T2I}\label{sec: revisiting rae}

The original RAE framework~\citep{zheng2025diffusion} introduced a suite of specialized design choices—including dimension-dependent noise scheduling, noise-augmented decoding, and a modified backbone ($\text{DiT}^{\text{DH}}$)—to enable diffusion on high-dimensional latents. While these modifications proved effective for class-conditional ImageNet generation, it remains unclear which are fundamental requirements for high-dimensional diffusion and which are adaptations for lower-capacity regimes.

In this section, we systematically stress-test these components under large-scale T2I settings. We systematically evaluate these components to determine their necessity in large-scale T2I generation. Our analysis reveals that adapting the noise schedule to the latent dimension is critical for convergence, whereas the architectural modifications proposed in the original work—such as wide diffusion heads and noise augmentation—become redundant at scale.

\subsection{Experiment Setup}\label{sec: rae experiment setup}

\paragraph{Model architecture.}
We adopt the MetaQuery architecture~\citep{metaquery} for text-to-image (T2I) generation and unified modeling. The model initializes with a pretrained language model (LLM) and prepends a sequence of learnable query tokens to the text prompt. The number of query tokens is set to 256, matching the number of visual tokens ($16 \times 16$) produced by the representation encoder. The LLM jointly processes the text and queries, producing query-token representations that serve as the conditioning signal. 
A 2-layer MLP connector then projects these representations from the LLM’s hidden space into the DiT model~\citep{dit}.

For this DiT model, we adopt the design based on LightningDiT~\citep{lgt} and train it using the flow matching objective~\citep{fm}. Critically, our model does not operate in a compressed VAE space. Instead, the DiT learns to model the distribution of high-dimensional, semantic representations generated by the frozen representation encoder. During inference, the DiT generates a set of features conditioned on the query tokens, which are then passed to our trained RAE decoder for rendering into pixel space. 

We also train visual instruction tuning~\citep{liu2023improved, liu2023visual} for image understanding. For this, we use a separate 2-layer MLP projector that maps visual tokens into the LLM’s embedding space. Importantly, these visual tokens come from the \textbf{same} frozen representation encoder whose features the diffusion model is trained to generate.

Unless otherwise specified, we use  SigLIP-2 So400M (patch size 14)~\citep{tschannen2025siglip} as our representation encoder and Qwen-2.5 1.5B~\citep{qwen2024qwen2} as the LLM in our experiments. We fix the number of visual tokens to 256, resulting in 224-resolution images for RAE and 256-resolution for VAE.

\paragraph{Flow matching.} Following standard practice, we adopt the flow matching objective~\citep{fm, rf} with linear interpolation $\mathbf{x}_t = (1 - t)\mathbf{x} + t \mathbf{\varepsilon}$, where $\mathbf{x} \sim p(\mathbf{x})$ and $\mathbf{\varepsilon} \sim \mathcal{N}(0, \mathbf{I})$, and train the model to predict the velocity $v(\mathbf{x}_t, t)$. Unless otherwise noted, we employ a 50-step Euler sampler for generation, consistent with RAE~\citep{zheng2025diffusion}.

\paragraph{Evaluation.} 
We evaluate using two widely adopted metrics: the GenEval score~\citep{ghosh2023geneval} and the DPG-Bench score~\citep{hu2024ella}.

\subsection{Noise scheduling remains crucial for T2I}
The RAE work~\citep{zheng2025diffusion} argues that conventional noise schedules become suboptimal when applied to high-dimensional latent spaces. 
The paper proposes a \emph{dimension-dependent noise schedule shift}~\citep{SD3} that rescales the diffusion timestep according to the effective data dimension $m = N \times d$ (number of tokens $\times$ token dimension). Formally, given a base schedule $t_n \in [0, 1]$ defined for a reference dimension $n$, the shifted timestep is computed as
\vspace{-0.3em}
\begin{equation*}
t_m = \frac{\alpha t_n}{1 + (\alpha - 1)t_n}, \quad \text{where} \quad \alpha = \sqrt{\frac{m}{n}}.
\end{equation*}
\vspace{-0.3em}

We follow the RAE setting and use $n{=}4096$ as the base dimension for computing the scaling factor $\alpha$.
We experiment with and without applying the dimension-dependent shift when training text-to-image diffusion models on RAE latents, as shown below. 

\begin{center}
{
\setlength{\tabcolsep}{6pt} 
\begin{tabular}{l|cc}
    \small
    \textbf{Setting} & \textbf{GenEval}$\uparrow$ & \textbf{DPG-Bench}$\uparrow$ \\
    \midrule
    w/o shift & 23.6 & 54.8 \\
    w/ shift    & \textbf{49.6} & \textbf{76.8} \\
\end{tabular}
}
\end{center}

Consistent with~\citet{zheng2025diffusion}, applying the noise shift dramatically improves both GenEval and DPG-Bench scores, demonstrating that adjusting the schedule to the effective latent dimension is critical for T2I.

\subsection{Design Choices that Saturate at Scale}\label{subsec: saturating designs}

While dimension-aware noise scheduling proves essential, we find that other design choices in RAE, which was originally developed for smaller-scale ImageNet models, provide diminishing returns at T2I scale.

\paragraph{Noise-augmented decoding.}
RAE originally proposed training decoders on perturbed latents to bridge the gap between training and inference distributions.
Formally, it trains the RAE decoder on smoothed inputs \(z' = z + n\), where \(n \sim \mathcal{N}(0,\, \sigma^2 I)\) and \(\sigma\) is sampled from \(|\mathcal{N}(0,\, \tau^2)|\).  We set $\tau = 0.2$ as we find a too high $\tau$ makes decoder training hard to converge.

\begin{figure}[t]
    \centering
    \begin{subfigure}[b]{0.49\textwidth}
        \centering
        \includegraphics[width=\textwidth]{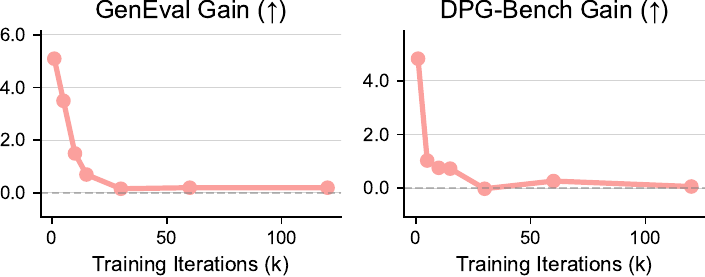}
        \vspace{-1em}
        \caption{
            Noise-augmented decoding gains diminish with training
        }\label{fig:noisy_decode_t2i}
    \end{subfigure}
    
    \vspace{0.75em}
    \begin{subfigure}[b]{0.49\textwidth}
        \centering
        \includegraphics[width=\textwidth]{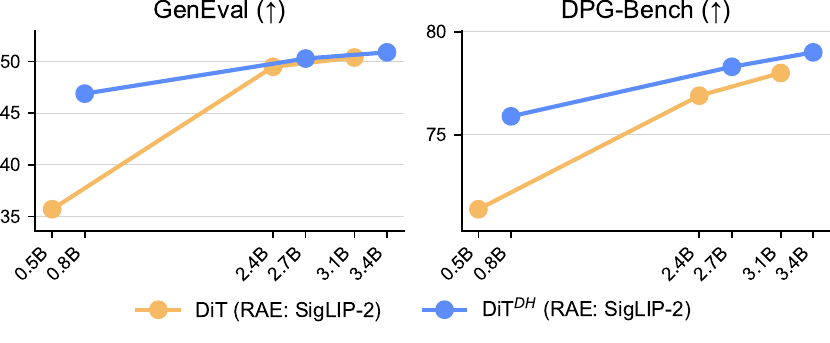}
        \vspace{-1.5em}
        \caption{\DDT{} advantage saturates as DiT scales}\label{fig:ditdh vs dit}
    \end{subfigure}
    \vspace{-1.5em}
    \caption{
        \textbf{Design choices that saturate at T2I scale.}
        \textit{Left}: Noise-augmented decoding provides substantial gains early in training but becomes negligible by 120k steps.
        \textit{Right}: \DDT{} yields large gains at 0.5B (+11.2 GenEval), but the advantage diminishes at $>$2.4B, where backbone capacity dominates.
    }\label{fig:saturating_designs}
    \vspace{-1em}
\end{figure}

We visualize the effect of noise-augmented decoding at different training stages in \cref{fig:noisy_decode_t2i}.
The gains are noticeable early in training (before \(\sim\)15k steps), when the model is still far from convergence, but become negligible at later stages.
This suggests that noise-augmented decoding acts as a form of regularization that matters most when the model has not yet learned a robust latent manifold.

\paragraph{Wide DDT head.}
The \DDT{} architecture augments a standard DiT with a shallow but wide DDT head, increasing denoising width without widening the entire backbone. In standard ImageNet-scale DiTs, the backbone width ($d \approx 1024$) is often narrower than the high-dimensional RAE latent targets ($d=1152$). \DDT{} circumvents this by appending a wide, shallow denoising head ($d=2688$) without incurring the cost of widening the full backbone.

However, T2I setting operates in a different regime. Modern large-scale T2I DiTs~\citep{flux, wu2025qwen} ($\geq$ 2B parameters) possess hidden dimensions ($d \geq 2048$) that inherently exceed the latent dimension. We hypothesize that this natural width eliminates the bottleneck \DDT{} was designed to fix.

To verify this, we train DiT variants across three scales---0.5B, 2.4B, and 3.1B---comparing standard architectures against counterparts augmented with the +0.28B parameter \DDT{} head. As shown in \cref{fig:ditdh vs dit}, the results confirm our hypothesis: at 0.5B, where the backbone is narrow, \DDT{} provides a critical +11.2 GenEval boost. Yet as the model scales to 2.4B and beyond, this advantage saturates greatly.

This finding clarifies that \DDT{} is a patch for capacity-constrained models, not a fundamental requirement for RAE. For scalable T2I training, standard DiT architectures are already sufficient.

\paragraph{Summary.}
Our experiments reveal clear design principles for scaling RAE-based diffusion models: dimension-aware noise scheduling remains non-negotiable, as it directly addresses the mathematical properties of high-dimensional latent spaces.
In contrast, architectural refinements (\DDT{}) and training augmentations (noise-augmented) that help at small scales provide diminishing returns as models grow---backbone capacity increasingly dominates performance.
From here on, we adopt standard DiT architectures with proper noise scheduling and no noise-augmented decoding.

\section{Training Diffusion Model with RAE vs. VAE}\label{sec: vae vs rae}

In this section, we compare text-to-image diffusion training using the RAE (SigLIP-2) encoder versus a standard VAE (FLUX-VAE). For the VAE baseline, we adopt the state-of-the-art model from FLUX~\citep{flux}. All experiments follow the same setup described in \cref{sec: rae experiment setup}, with identical training configurations; the only difference lies in whether diffusion is performed in the RAE or VAE latent space. We defer implementation details to \cref{appendix: implementation}.

\paragraph{Experimental Protocol.}
We organize our comparison into two stages: \emph{pretraining} and \emph{finetuning}. We train the Diffusion Transformer \emph{from scratch} in both settings to ensure a fair comparison of convergence speed and data efficiency. We ensure apples-to-apples comparison. The \emph{only} component that differs is the latent space and its decoder (SigLIP-2 RAE vs.\ FLUX VAE). 
For the VAE baseline, we employ FLUX VAE for generation while retaining the SigLIP encoder for understanding, as VAE latents are insufficient for perception~\citep{zhou2024transfusion}. This design choice effectively forms a two-tower architecture, mirroring the design of recent unified models like Bagel~\citep{deng2025emerging} and UniFluid~\citep{fan2025unified}.

\paragraph{Pretrain Data.}
We follow the data mixture developed in FuseDiT~\citep{Tang_2025_CVPR} and adopt the recaptioned texts and remixing ratios released by BLIP-3o~\citep{blip3o}. The mixture combines mostly webdata like CC12M~\citep{changpinyo2021conceptual}, SA-1B~\citep{kirillov2023segment}, and JourneyDB~\citep{sun2023journeydb}, totaling approximately 39.3M images. In addition, we use FLUX.1-schnell~\citep{flux} to generate 24.7M synthetic images.
We also train on Cambrian-7M~\citep{tong2024cambrian} to develop the model's visual understanding capabilities.

\begin{table}[t]
    \centering
    \small
    \caption{
        \textbf{Data composition matters more than scale.}
        Synthetic data substantially outperforms web data, and their combination (49.5 GenEval) surpasses even doubled synthetic data (48.0), demonstrating synergistic benefits from complementary data sources rather than volume alone.
    }\label{tab:data_recipe}
    \begin{tabular}{lcc}
        \toprule
        \textbf{Training Data} & \textbf{GenEval} $\uparrow$ & \textbf{DPG-Bench} $\uparrow$ \\
        \midrule
        Synthetic & 45.1 & 73.8 \\
        Synthetic $\times$2 & 48.0 & 75.2 \\
        \midrule
        Web & 25.9 & 69.5 \\
        Web $\times$2 & 26.3 & 70.6 \\
        \midrule
        \rowcolor{green!10}
        Synthetic + Web & \textbf{49.5} & \textbf{76.9} \\
        \bottomrule
    \end{tabular}
    \vspace{-0.5em}
\end{table}

We experiment with a Qwen-2.5 1.5B LLM and a 2.4B DiT to study how different pretraining corpora influence text-to-image performance. 
We train three variants: (i) Web-39M + Cambrian-7M, (ii) FLUX-generated synthetic data + Cambrian-7M, and (iii) their union. 
As shown in \cref{tab:data_recipe}, the mixed dataset yields the best performance.

To ensure the gains are not simply due to more data, we also double the size of each individual source (Web ×2, Synthetic ×2). 
These runs yield much smaller improvements, indicating that the benefits arise from the complementary nature of the two data types rather than data volume alone.

We also find that synthetic data results in lower training loss and faster convergence, suggesting that FLUX images provide more stylistically consistent signals. 
Web-scale data, by contrast, is harder to fit but provides more diverse signals. 
When combined, the model inherits visual style from synthetic data and rich semantics from web data, leading to clear and robust improvements in generation quality.

\subsection{Pretraining}\label{sec: pretraining}

\paragraph{Convergence.}
We first compare the convergence behavior.
We train a Qwen2.5-1.5B LLM with a 2.4B DiT backbone.
As shown in \cref{fig:convergence}, the RAE-based model converges significantly faster than its VAE counterpart, achieving a 4.0× speedup on GenEval and a 4.6× speedup on DPG-Bench.

\paragraph{Scaling DiT models.}

\begin{figure}[t]
    \centering
    \begin{subfigure}[b]{0.49\textwidth}
        \centering
        \includegraphics[width=\textwidth]{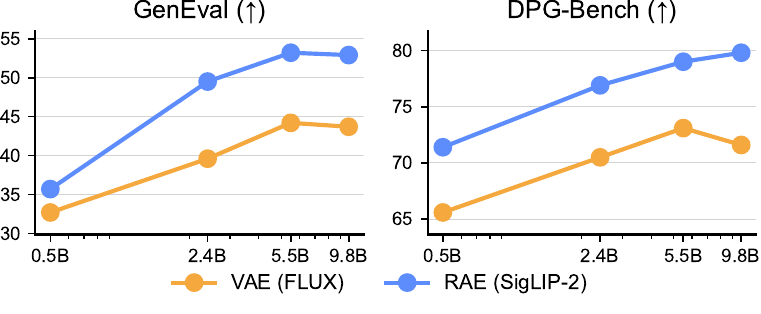}
        \vspace{-1.5em}
        \caption{Scaling DiT models with fixed LLM (Qwen2.5 1.5B)}\label{fig:scale dit}
        \vspace{1em}
    \end{subfigure}
    \hfill
    \begin{subfigure}[b]{0.49\textwidth}
        \centering
        \includegraphics[width=\textwidth]{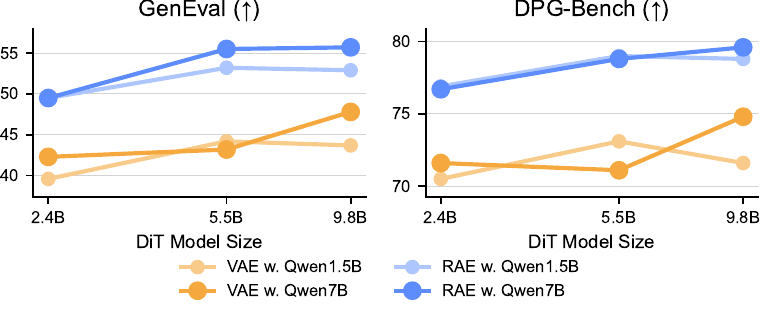}
        \vspace{-1.5em}
        \caption{Scaling LLM and DiT jointly}\label{fig:scale llm}
        \vspace{1em}
    \end{subfigure}
    \vspace{-2.5em}
    \caption{
\textbf{RAE outperforms VAE across LLM and DiT scales.} 
\textit{Top}: With a 1.5B LLM, RAE-based models outperform VAE-based ones at all DiT sizes (0.5B, 2.4B, 5.5B, 9.8B). 
\textit{Bottom}: Using a larger 7B LLM, RAE continues to maintain its advantage.
    }\label{fig:scaling}
    \vspace{-1.25em}
\end{figure}

We use Qwen-2.5 1.5B as the language backbone, and train DiT variants of 0.5B, 2.4B, 5.5B, and 9.8B parameters. The architectures of these DiT variants are designed following recent advances in large-scale vision models~\citep{esser2024scaling, fan2025scaling, wan2025wan, wu2025qwen}, and detailed model specifications are provided in \cref{appendix: model}. In this experiment, we train all the models for 30k iterations with a batch size of 2048. 

In \cref{fig:scale dit}, we find that RAE-based models consistently outperform their VAE counterparts at all scales. 
Even for the smallest 0.5B DiT, where the network width only slightly exceeds the RAE latent dimension, the RAE-based model still shows clear advantages over the VAE baseline.

We also observe diminishing returns when scaling DiT models beyond 6B parameters. The performance trend appears to plateau, suggesting that simply increasing model size without proportionally improving data quality and diversity may lead to underutilized capacity.
This observation aligns with discussions in large-scale visual SSL literature~\citep{fan2025scaling}, which highlight the need for high-quality data scaling to fully exploit model capacity.

\paragraph{Scaling LLM backbones.}
We study how scaling the LLM backbone influences text-to-image performance when paired with diffusion models of different sizes. 
To this end, we train both RAE- and VAE-based models using LLMs of \{1.5B, 7B\} parameters combined with DiTs of \{2.4B, 5.5B, 9.8B\}, and present the results in~\cref{fig:scale llm}.

We observe performance gains from scaling the LLM to 7B, particularly when paired with RAE. We note that prior studies, such as MetaQuery~\citep{metaquery}, reported limited benefits from LLM scaling. Our results diverge from this conclusion, likely due to two key factors: (1) we evaluate on significantly larger diffusion backbones (up to 9.8B) which can better exploit rich text representations, and (2) we finetune the LLM backbone, allowing it to adapt its latent space for generative tasks more effectively than frozen approaches.

\paragraph{Generalizing to other vision encoders.}
We also experiment with training RAE with WebSSL ViT-L~\citep{fan2025scaling}. 
Under the same 1.5B LLM and 2.4B DiT setup, the WebSSL RAE performs slightly below the SigLIP-2 version but still exceeds the FLUX VAE baseline (\cref{tab:ssl enc}). This finding is notable because WebSSL is not explicitly aligned with text; it suggests that the RAE framework in T2I training is robust to the choice of encoder.

\begin{table}[t]
    \centering
    \small
    \caption{
    \textbf{SSL encoders are effective RAE backbones for T2I.} 
    A WebSSL-based RAE performs slightly worse than SigLIP-2 but remains stronger than FLUX VAE in T2I.
    }
    \label{tab:ssl enc}
    \begin{tabular}{lcc}
        \toprule
        \textbf{Model Variant} & \textbf{GenEval} $\uparrow$ & \textbf{DPG-Bench} $\uparrow$ \\
        \midrule

        \rowcolor{gray!10}
        \multicolumn{3}{l}{\textbf{VAE-based models}} \\
        FLUX VAE & 39.6 & 70.5 \\

        \rowcolor{gray!10}
        \multicolumn{3}{l}{\textbf{RAE-based models}} \\
        WebSSL ViT-L & 46.0 & 72.8 \\
        SigLIP-2 ViT-So & 49.5 & 76.9 \\
        
        \bottomrule
    \end{tabular}
    \vspace{-1em}
\end{table}

\subsection{Finetuning}
\label{sec: finetune}
Following standard practice in T2I training~\citep{dai2023emu, pixartalpha, podell2023sdxl}, models are finetuned on a smaller high-quality dataset after large-scale pretraining. 
We evaluate this finetuning stage for both RAE- and VAE-based models under identical settings. 
Unless otherwise noted, we use the BLIP-3o 60k dataset~\citep{blip3o} and start from the 1.5B LLM + 2.4B DiT checkpoint trained for 30k steps in \cref{sec: pretraining}. We update both the LLM and the DiT; additional details are provided in \cref{appendix: implementation}.

\begin{figure}[t]
    \centering
    \includegraphics[width=\columnwidth]{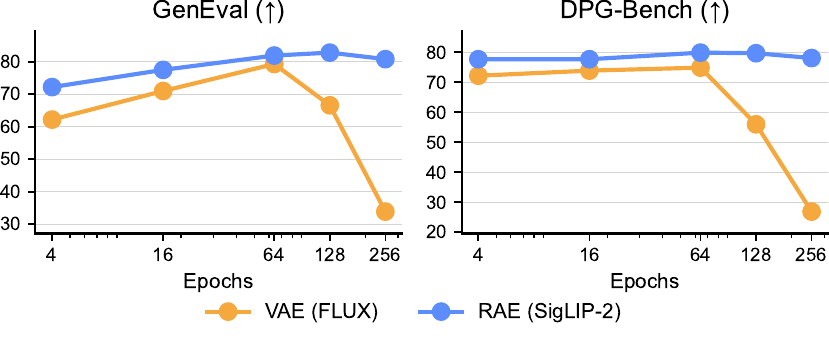}
    \caption{
        \textbf{RAE-based models outperform VAE-based models and are less prone to overfitting.} 
        We train both models for 256 epochs and observe that (1) RAE-based models consistently achieve higher performance, and (2) VAE-based models begin to overfit rapidly after 64 epochs.
    }\label{fig:finetune comparison}
\end{figure}

\paragraph{RAE-based models consistently outperform VAE-based models.} We finetune both family of models for \{4, 16, 64, 128, 256\} epochs and compare the performance on GenEval and DPG-Bench in \cref{fig:finetune comparison}. 
We observe that across all iterations, the RAE-based model shows an advantage on both GenEval and DPG-Bench across all settings. 

\paragraph{RAE-based models are less prone to overfitting.}
As shown in \cref{fig:finetune comparison}, VAE-based models degrade significantly after 64 epochs. Training loss analysis (Appendix \cref{fig:finetune loss}) reveals that the VAE loss collapses rapidly to near-zero, suggesting the model is memorizing individual training samples rather than learning the underlying distribution. In constrast, RAE-based models remain stable and show only a mild decline. We hypothesize that the higher-dimensional and semantically structured latent space of the RAE\footnote{SigLIP-2 produces 1152-dim.\ tokens vs.\ $<$100 in typical VAEs}
may provide an implicit regularization effect, helping mitigate overfitting during finetuning.

\begin{figure}[t]
    \centering
    \includegraphics[width=\columnwidth]{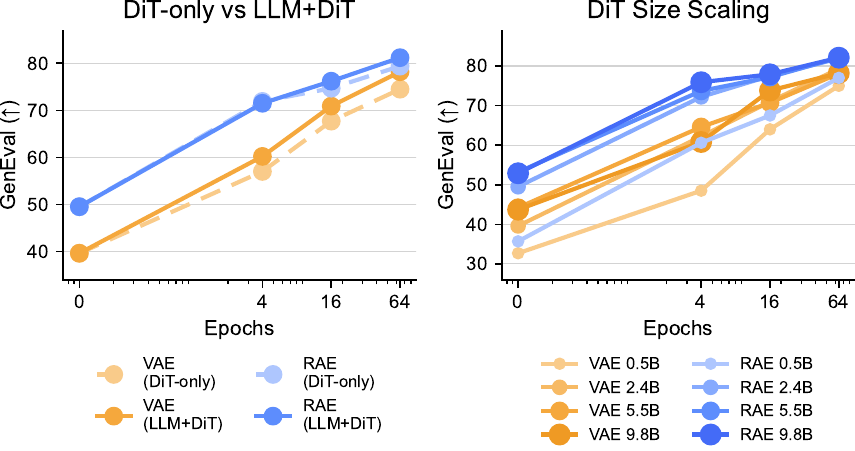}
    \vspace{-2em}
    \caption{
        \textbf{RAE-based models outperform VAEs across different settings.}
        \textit{Left}: When fine-tuning only the DiT versus the full LLM+DiT system, RAE models consistently achieve higher GenEval scores.
        \textit{Right}: RAE models maintain their advantage over VAE across all DiT model scales (0.5B--9.8B parameters), with the performance gap widening as model size increases.
    }\label{fig:posttrain ablation}
    \vspace{-1em}
\end{figure}

\paragraph{RAE's advantage generalizes across settings.}
To verify whether RAE's advantage over VAE extends beyond our main setup, we conduct two additional experiments: 1) fine-tuning only the DiT while freezing the LLM (following recent works~\cite{metaquery, blip3o}), and 2) scaling to different sizes DiT models (0.5B--9.8B parameters).
\Cref{fig:posttrain ablation} shows that RAE consistently outperforms VAE in both settings. The left panel shows that both selective fine-tuning (DiT-only) and joint fine-tuning (LLM+DiT) favor RAE over VAE; notably, the top-performing VAE configuration reaches 78.2, while the weakest RAE approach achieves 79.4. The right panel shows continued RAE gains across the scaling range, with larger models exhibiting greater improvements.

\section{Implications for Unified Models}\label{sec: unified model implications}

A key advantage of the RAE framework is that it unifies visual understanding and generation within a \emph{single, shared, high-dimensional latent space}. This contrasts with the conventional two-tower design (used in our Section~\ref{sec: vae vs rae} baseline). In two-tower models, the generation head operates in a latent space alien to the LLM's understanding encoder. This effectively makes the unified model 'blind' to its own output distribution without a VAE decoder. In contrast, RAE forces generation to occur in the \emph{same representation space} of the visual encoder. This means the model generates the exact same high-dimensional features it uses to see.

\begin{figure}[t]
    \centering
    \includegraphics[width=\columnwidth]{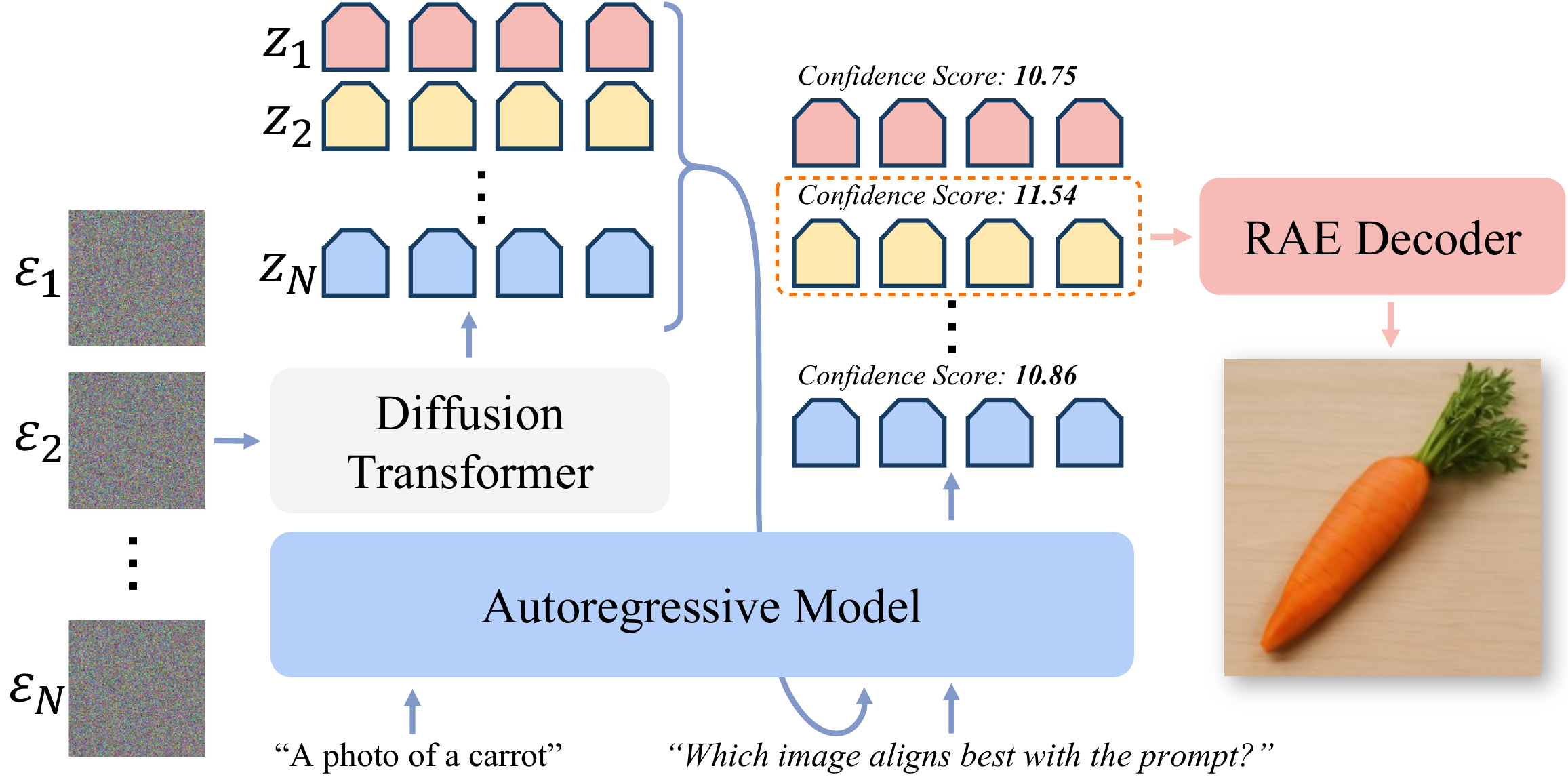}
    \caption{
        \textbf{Test-time scaling in latent space.}
        Our framework allows the LLM to directly evaluate and select generation results within the latent space, bypassing the decode-re-encode process.
    }\label{fig:tts}
    \vspace{-1em}
\end{figure}

\paragraph{Test-time scaling in latent space.}
A direct benefit of this shared representation is that the LLM can interpret the latents produced by the diffusion model without needing to decode them into pixels and re-encode them, leaving the representation and pixel spaces fully decoupled. We leverage this property to implement \textbf{Latent Test-Time Scaling (TTS)}, where the LLM acts as a verifier for its own generations directly and only in the feature space (\cref{fig:tts}).

We explore two verifier metrics that leverage the LLM's understanding capabilities to score generated latents:
(1) \textbf{Prompt Confidence}: We re-inject the generated latents and the original text prompt back into the LLM and measure the aggregate token-level confidence of the prompt, following~\citet{kang2025scalable}.
(2) \textbf{Answer Logits}: We explicitly query the LLM with: \textit{``Does this generated image $\langle$image$\rangle$ align with the $\langle$prompt$\rangle$?''} and use the logit probability of the ``Yes'' token as the score.

With the verifier defined, we adopt the standard test-time scaling protocol~\citep{ma2025inference, xie2025sana} using a best-of-$N$ selection strategy. As shown in~\cref{tab:tts}, both verification metrics yield consistent improvements on GenEval, demonstrating that latent-space TTS is not only feasible but also an effective way to enhance generation quality.
Crucially, this improvement is achieved entirely within the semantic latent space, demonstrating that the model can verify the quality of its own generations without ever needing to render pixels.

\begin{table}[t]
    \centering
    \small
    \setlength{\tabcolsep}{5pt}
    \caption{\textbf{TTS results across LLM–DiT configurations.} Substantial performance improvements observed with both verifier metrics on GenEval. 4/8 refers to ``selecting best 4 out of 8''.}
    \label{tab:tts}
    \begin{tabular}{lccc}
        \toprule
        \textbf{Best-of-$N$} & \textbf{Prompt Confidence} & \textbf{Answer Logits} \\
        \midrule
        \rowcolor{gray!10}
        \multicolumn{3}{l}{\scriptsize \textbf{1.5B LLM + 5.5B DiT \quad (GenEval = 53.2)}} \\
        4/8      & 56.7  & 59.6   \\
        4/16      & 57.5  & 62.5 \\
        4/32      & 60.0  & 64.3  \\
        
        \rowcolor{gray!10}
        \multicolumn{3}{l}{\scriptsize  \textbf{7.0B LLM + 5.5B DiT \quad (GenEval = 55.5)}} \\
        4/8     & 58.3  & 62.5\\
        4/16    & 59.6  & 65.8   \\
        4/32    & 60.1  & 67.8  \\
        \bottomrule
    \end{tabular}
    \vspace{-1em}
\end{table}

\paragraph{Visual understanding.}
Finally, we conduct a comparative study to study how the choice of visual generation backbone—VAE versus RAE—affects multimodal understanding performance. 
We evaluate the trained models on standard benchmarks: MME~\citep{fu2023mme}, TextVQA~\citep{singh2019towards}, AI2D~\citep{hiippala2021ai2d}, SeedBench~\citep{ge2023planting}, MMMU~\citep{yue2023mmmu}, and MMMU-Pro~\citep{yue2024mmmu}. 
We emphasize that the goal of this work is not to build a SOTA VQA model; achieving that would require additional components such as any-resolution inputs, multimodal continual pretraining, and very high-quality data.

Similar to prior findings~\citep{metamorph,fan2025unified,deng2025emerging}, we observe in \cref{tab:vl_eval} that adding generative modeling does not degrade visual understanding performance. The choice of RAE vs.\ VAE in the generative path has little impact, likely because both variants share the same frozen understanding encoder.

\begin{table}[t]
    \centering
    \footnotesize
    \setlength{\tabcolsep}{4.2pt}  
    \caption{
\textbf{Generative training leaves understanding intact; RAE and VAE perform similarly.}
Across VL benchmarks, both latent choices produce comparable understanding performance.
}\label{tab:vl_eval}
    \begin{tabular}{lcccccc}
        \toprule
        \textbf{Model} & 
        \textbf{MME$_{\text{P}}$} &
        \textbf{TVQA} &
        \textbf{AI2D} &
        \textbf{Seed} &
        \textbf{MMMU} &
        \textbf{MMMU$_{\text{P}}$} \\
        \midrule
        Und.-only  &  1374.8 &  44.7 & 63.9 & 67.1 & 40.2 & 20.5 \\
        RAE-based  &  1468.7 &  39.6 & 66.7 & 69.8 & 41.1 & 19.8 \\
        VAE-based  &  1481.7 &  39.3 & 66.7 & 69.7 & 37.2 & 18.7 \\
        \bottomrule
    \end{tabular}
\end{table}

\section{Related Work}

\paragraph{VAE, representation and representation autoencoder.}

A common line of work uses VAEs~\citep{VAE} as autoencoders to compress images into low-dimensional latent spaces, typically with channel dimensions below 64~\citep{flux, SD3}. 
Many studies~\citep{dcae,titok} have pursued aggressive compression, while others~\citep{vqvae,vqvae2} reduce dimensionality further by quantizing continuous latents into discrete codes. However, both directions unavoidably result in information loss.

Representation Autoencoders (RAE)~\citep{zheng2025diffusion} take a different route: use a frozen, pretrained representation encoder and train only the decoder to reconstruct images from high-dimensional semantic features. 
In ImageNet experiments, training diffusion transformers~\citep{dit} in this latent space yields faster convergence and better performance than VAEs. In this work, we extend RAE to text-to-image generation and show that its reconstruction and generative advantages transfer naturally to the multimodal setting. 

Recently, several works have explored leveraging representation encoders for reconstruction. SVG~\citep{svg, svg-t2i} employs a residual encoder to refine visual details during reconstruction, while VTP~\citep{VTP} incorporates a reconstruction loss into the pretraining of representation encoders. VQRAE~\citep{vqrae} further applies quantization on top of representation encoders to construct discrete representations for generation. In a related direction, another line of work~\citep{reglue,redi,semvae} integrates representation encoders with VAEs to improve generation fidelity.

\paragraph{VAE in Text-to-image models.}
VAE has also been widely used in text-to-image models. Stable Diffusion~\citep{LDM} uses an off-the-shelf VAE and a text-conditioned U-Net~\citep{unet} for T2I training. Subsequent work~\citep{podell2023sdxl, SD3,flux} improves VAE through higher-quality and larger-scale training data. 

Recently, Stable Diffusion 3~\citep{SD3} shows that widening the VAE channels boosts reconstruction fidelity and enhances the scalability of the downstream diffusion model, while Hunyuan-Image-3~\citep{cao2025hunyuanimage} further incorporates representation alignment~\citep{repa} into VAE training.

This work takes the representation route a step further: instead of modifying VAEs, we train T2I models directly on high-dimensional representation spaces with RAE. This approach yields clear advantages over VAE in both convergence speed and final generation quality.

\paragraph{Unified Multimodal Models.} 
Recently, many works focus on unifying multimodal understanding and generation into one modeling paradigm. 
One stream of work discretizes visual input and trains next token prediction modeling~\cite{team2024chameleon,wang2024emu3,chen2025janus, wu2024vila,jiao2025unitoken}.
Another stream of research incorporates diffusion model into LLMs~\cite{dai2023emu,dong2023dreamllm,emu2,ge2024seed,metamorph,zhou2024transfusion, metaquery,blip3o,deng2025emerging}.  However, it has been viewed that understanding and generation require \emph{different} visual representations---high-dimensional CLIP features for understanding and low-dimensional VAE latents for generation---leading most unified models to adopt a two-tower design.

An emerging direction in unified multimodal modeling is to unify \emph{understanding} and \emph{generation} into a shared latent space. 
To work around this mismatch, recent approaches~\citep{unilip,vugen,uniflow,atoken,huang2025ming} adopt continuous representation spaces but introduce substantial downsampling for generation. 
For example, \citet{vugen} uses high-dimensional, uncompressed features for understanding but falls back to compressed, lower-dimensional latents for generation. 
\citet{jiao2025unitoken} and \citet{uniflow} employ compressed embeddings for both understanding and generation
, limiting the model's perception ability. 
BLIP-3o~\citep{blip3o} experiments with using a Qwen2.5-VL encoder~\citep{bai2025qwen2} for understanding and EVA-CLIP~\citep{sun2023eva, sun2024eva} for generation; However, because the model does not apply noise-schedule shifting and its DiT width is smaller than the EVA-CLIP embedding dimension, it relies on a strong diffusion decoder~\citep{podell2023sdxl} to map these features back to pixels.

Our work takes a step forward by using a \emph{single high-dimensional encoder} for both understanding and generation. 
Leveraging RAE designs, the model enjoys a simpler architecture that understands and generates directly in this semantic space, surpassing VAE-based designs in T2I.

\section{Conclusion}
In this work, we investigate scaling Representation Autoencoders (RAEs) to text-to-image generation. Our study begins by scaling the decoder, where we find that while larger data scales improve general fidelity, specific domains such as text require targeted data composition. We then examine the RAE framework itself, revealing that scaling simplifies the design: dimension-dependent noise scheduling remains essential, but architectural modifications like \DDT{} yield diminishing returns as model capacity increases. Building on this streamlined recipe, we show that RAE-based diffusion models consistently outperform state-of-the-art VAE baselines in convergence speed and generation quality, while being less prone to overfitting during finetuning. Collectively, these results establish RAE as a simple and effective foundation for large-scale generation. Moreover, by enabling understanding and generation to operate in a shared representation space, RAEs open new possibilities for unified models, such as the latent-space test-time scaling demonstrated in this work. We believe RAE serve as a strong foundation for future research in both scalable generation and unified multimodal modeling.

\section{Acknowledgements}
The authors would like to thank Xichen Pan, Shusheng Yang, David Fan, John Nguyen for insightful discussions and
feedback on the manuscript. This work was mainly supported by the Google TPU Research Cloud
(TRC) program and the Open Path AI Foundation. ST is supported by Meta AI Mentorship Program.
SX also acknowledges support from the MSIT IITP grant (RS-2024-00457882) and the NSF award
IIS-2443404.

\clearpage
{
    \small
    \bibliographystyle{ieeenat_fullname}
    \bibliography{main}
}

\clearpage

\appendix
\begin{table*}[t!]
\centering

\begin{subtable}{0.48\textwidth}
\centering
\small
\begin{tabular}{lll}
\toprule
\textbf{Component} & \textbf{Decoder} & \textbf{Discriminator} \\
\midrule
optimizer                  & AdamW               & AdamW \\
max learning rate         & $4 \times 10^{-4}$  & $5 \times 10^{-5}$ \\
min learning rate         & $4 \times 10^{-5}$  & $5 \times 10^{-6}$\\
learning rate schedule     & cosine decay        & cosine decay \\
optimizer betas            & (0.9, 0.95)          & (0.9, 0.95) \\
weight decay               & 0.0                 & 0.0 \\
batch size                 & 512                 & 512 \\
warmup                     & 2 epoch            & 1 epoch \\
loss                       &{\scriptsize $\ell_1$ + LPIPS + GAN + Gram} & adv. \\
Model                       & ViT-XL& Dino-S/16 (frozen) \\
LPIPS start epoch          & 0 & -- \\
disc. start epoch    & -- &7 \\
adv. loss start epoch    & 8 & --\\
Training epochs           & 80 & 73 \\
\bottomrule
\end{tabular}
\end{subtable}
\hfill
\begin{subtable}{0.48\textwidth}
\centering
\small
\begin{tabular}{lll}
\toprule
\textbf{Component} & \textbf{Decoder} & \textbf{Discriminator} \\
\midrule
optimizer                  & AdamW               & AdamW \\
max learning rate         & $2 \times 10^{-4}$  & $2 \times 10^{-5}$ \\
min learning rate         & $2 \times 10^{-5}$  & $2 \times 10^{-6}$\\
learning rate schedule     & cosine decay        & cosine decay \\
optimizer betas            & (0.9, 0.95)          & (0.9, 0.95) \\
weight decay               & 0.0                 & 0.0 \\
batch size                 & 512                 & 512 \\
warmup                     & 2 epoch            & 1 epoch \\
loss                       & {\scriptsize $\ell_1$ + LPIPS + GAN + Gram} & adv. \\
Model                       & ViT-XL& Dino-S/16 (frozen) \\
LPIPS start epoch          & 0 & -- \\
disc. start epoch    & -- &10 \\
adv. loss start epoch    & 11 & --\\
Training epochs           & 80 & 70 \\
\bottomrule
\end{tabular}
\end{subtable}
\caption{\textbf{Training configuration for decoder and discriminator.}
\textit{Left}: Configuration used for SigLIP2-So.
\textit{Right}: Configuration used for WebSSL ViT-L.
Different encoders require slightly different training recipes for achieving strong decoder performance. }
\label{tab:recon_training_config}
\end{table*}
\section{Implementation}\label{appendix: implementation}
Our experiments are conducted on TPU v4, v5p, and v6e with TorchXLA.
\paragraph{Decoder training.} We largely follow RAE for decoder architecture and adopt ViT-XL~\citep{vit} as the default decoder. The decoder contains 28 blocks with a hidden size of 1152 and 16 heads.
Decoder training details are included in Table~\ref{tab:recon_training_config}. We find the GAN training recipe provided in~\citep{zheng2025diffusion} is not stable on web-scale images. To tackle the issue, we tune the recipe to as~\cref{tab:recon_training_config}. On web-scale images, we find using DINO-S/16 already suffices as a strong discriminator, and using DINO-S/8 as in~\citep{zheng2025diffusion} makes it hard to converge. Therefore, we use DINO-S/16 as the default discriminator. All input is interpolated to $224 \times 224$ resolution before feeding into the discriminator. We use an epoch-based training scheme and set the sample of each virtual epoch to be the same as ImageNet (1.28M). For loss coefficients, we set $\omega_G = 100.0, \omega_L = 1.0, \omega_A = 10.0$.

\paragraph{T2I \& unified model pretraining.}

For pretraining experiments in \cref{sec: pretraining}, we primarily train on TPU-v5p-128 and TPU-v6e-64. Detailed training configurations are provided in \cref{tab: pretrain configs}. We find that finetuning a pretrained LLM while training the DiT from scratch benefits from using separate optimizers, and properly decoupling their optimizer settings substantially improves training stability. We use SPMD sharding~\citep{barham2022pathways} together with TorchXLA to train the LLM, adapters, and DiT models.

\begin{table}[h]
\centering
\small
\setlength{\tabcolsep}{3.4pt}  
\begin{tabular}{lcc}
\toprule
\textbf{Component}        & \textbf{LLM}                    & \textbf{DiT} \\ 
\midrule
optimizer                 & \multicolumn{2}{c}{AdamW} \\
learning rate schedule    & \multicolumn{2}{c}{cosine w/ warmup ratio $0.0134$} \\
global batch size         & \multicolumn{2}{c}{$2048$} \\
\midrule
max learning rate         & $5 \times 10^{-5}$              & $5 \times 10^{-4}$ \\
optimizer betas           & $(0.9,\ 0.999)$                 & $(0.9,\ 0.95)$ \\
loss                      & autoregressive LM               & diffusion loss \\
model                     & Qwen2.5 (1.5B / 7B)             & DiT (0.5B–9.8B) \\
\bottomrule
\end{tabular}
\caption{\textbf{Optimization hyperparameters for the LLM backbone and the DiT diffusion head in the unified T2I model.}}
\label{tab: pretrain configs}
\end{table}

\paragraph{T2I \& unified model finetuning.}
We finetune pre-trained models on the BLIP3o-60k dataset~\cite{chen2025blip3} using TPU-v4-128 and TPU-v5p-64. To ensure a fair comparison, we apply identical training configurations to both RAE and VAE models across 4, 16, 64, and 256 epochs. We utilize the same codebase and training infrastructure as the pretraining stage. We use a global batch size of 1024 and the optimization settings detailed in Tab.~\ref{tab:finetune_hyperparams}.

\begin{table}[h]
\centering
\small
\setlength{\tabcolsep}{3.4pt}  
\begin{tabular}{lcc}
\toprule
\textbf{Component}        & \textbf{LLM}                    & \textbf{DiT} \\ 
\midrule
optimizer                 & \multicolumn{2}{c}{AdamW} \\
learning rate schedule    & \multicolumn{2}{c}{cosine w/ warmup ratio $0.03$} \\
global batch size         & \multicolumn{2}{c}{$1024$} \\
training epochs           & \multicolumn{2}{c}{4, 16, 64, 256} \\
\midrule
max learning rate         & $5.66 \times 10^{-5}$               & $5.66 \times 10^{-4}$ \\
optimizer betas           & $(0.9,\ 0.999)$                 & $(0.9,\ 0.95)$ \\
loss                      & autoregressive LM               & diffusion loss \\
model                     & Qwen2.5 (1.5B / 7B)             & DiT (0.5B–9.8B) \\
\bottomrule
\end{tabular}
\caption{\textbf{Finetuning hyperparameters.}}
\label{tab:finetune_hyperparams}
\end{table}

\paragraph{Synthetic data generation.}
For synthetic image generation, we compile prompts from publicly available prompt datasets\footnote{\url{https://huggingface.co/datasets/Geonmo/midjourney-prompts-only},
\url{https://huggingface.co/datasets/FredZhang7/stable-diffusion-prompts-2.47M},
\url{https://huggingface.co/datasets/neuralworm/stable-diffusion-discord-prompts},
\url{https://huggingface.co/datasets/isidentical/random-stable-diffusion-prompts},
\url{https://huggingface.co/datasets/CaptionEmporium/coyo-hd-11m-llavanext}}. 
Using these prompts, we generate 24.7M synthetic images with FLUX.1-schnell~\citep{flux}, which form part of our decoder training and T2I training corpus. We perform large-scale generation using TPU-v6e and will open-source the inference pipeline to facilitate future research.

\section{Models}\label{appendix: model}
\paragraph{LLM model and unified model configs.} We use pretrained Qwen2.5~\citep{qwen2024qwen2} language models at the 1.5B and 7B scales in our experiments. Following prior work~\citep{liu2023visual,tong2024cambrian}, we use a 2-layer MLP to project visual features from the representation encoder into the LLM embedding space, and a separate linear layer to map the LLM’s query-token outputs into the input space of the diffusion model.

\paragraph{DiT Model configs.}
We design our diffusion architecture following LightningDiT~\citep{lgt}. Motivated by recent findings in scaling vision backbones~\citep{fan2025scaling, wan2025wan, dehghani2023scaling}, we prioritize increasing model \emph{width} rather than depth when scaling DiT models. Consistent with insights from the RAE paper~\citep{zheng2025diffusion}, we also ensure that the DiT hidden dimension remains strictly larger than the target latent dimension (e.g. 1152 for the SigLIP2 ViT-So model), including at small scales such as DiT-0.5B. The detailed model specifications are provided in \cref{tab:diffusion model spec}.

\begin{table}[h]
\centering
\small
\begin{tabular}{lccc}
\toprule
Model & Hidden size & Heads & Depth \\
\midrule
DiT-0.5B & 1280  & 32 & 16 \\
DiT-2.4B & 2048  & 32 & 32 \\
DiT-3.3B & 2304 & 32 & 32 \\
DiT-5.5B & 3072  & 32 & 32 \\
DiT-9.8B & 4096  & 32 & 32 \\
\bottomrule
\end{tabular}
\caption{\textbf{Architectural specifications of DiT variants.}}
\label{tab:diffusion model spec}
\end{table}

\section{Additional Results}\label{appendix: additional results}

\begin{figure}[h]
    \centering
    \includegraphics[width=\columnwidth]{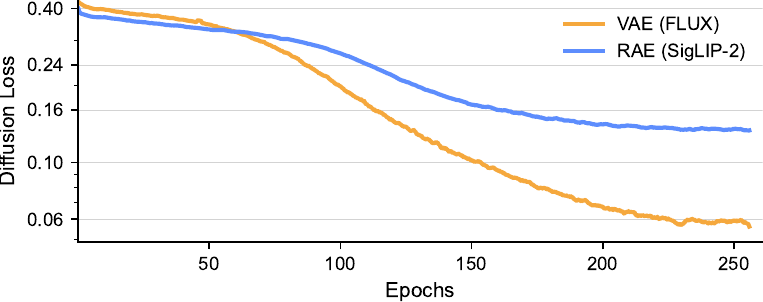}
    \vspace{-1.75em}
    \caption{
        \textbf{Diffusion loss during finetuning (256 epochs).}
        RAE overfits less and later than VAE: the VAE loss plunges early to very low values, while the RAE loss decreases more gradually and plateaus at higher values, indicating reduced overfitting.
    }\label{fig:finetune loss}
\end{figure}

\paragraph{Training losses.}
To complement the results in \cref{sec: finetune}, we additionally compare the training loss curves of RAE and VAE models during finetuning in \cref{fig:finetune loss}. We observe that the VAE model's loss decreases rapidly to a very low value, which correlates with the performance degradation observed in \cref{fig:finetune comparison}, a clear sign of overfitting. In contrast, the RAE model's loss decreases more gradually and stabilizes at a higher value, maintaining robust generation performance throughout the training process. This suggests that the high-dimensional semantic space of RAE provides a form of implicit regularization that prevents the model from memorizing the small finetuning dataset.

\begin{figure}[h]
    \centering
    \includegraphics[width=0.7\columnwidth]{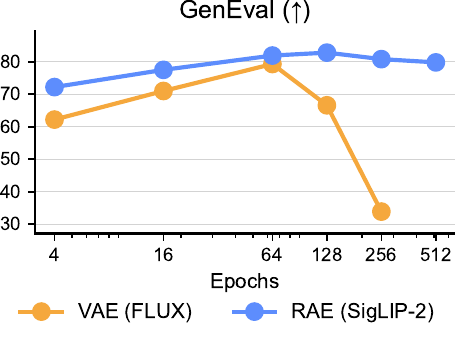} 
    \vspace{-1em}
    \caption{
        \textbf{Extended finetuning to 512 epochs.} 
        RAE maintains robust performance even with 512 epochs of training, while VAE suffers catastrophic overfitting after 64 epochs.
    }\label{fig:rae_vae_2_4b_epochs}
    \vspace{-1em}
\end{figure}

\paragraph{Extending finetuning to 512 epochs.}
We extend the finetuning experiment from \cref{sec: finetune} to 512 epochs. As shown in \cref{fig:finetune comparison}, VAE-based models already suffer substantial performance drops by 256 epochs, so we do not continue training them further. In contrast, the RAE-based model remains stable: even after 512 epochs (\cref{fig:rae_vae_2_4b_epochs}), it shows only a small decline in performance. This further supports the robustness of RAE-based methods under long-horizon finetuning.

\end{document}